\documentclass{article}

\usepackage[preprint]{neurips_2024}
\makeatletter
\renewcommand{\@noticestring}{This work won the OpenAI GPT-OSS-20B Red Teaming Challenge on Kaggle}
\makeatother

\usepackage[utf8]{inputenc} 
\usepackage[T1]{fontenc}    
\usepackage{hyperref}       
\usepackage{url}            
\usepackage{booktabs}       
\usepackage{amsfonts}       
\usepackage{nicefrac}       
\usepackage{microtype}      
\usepackage{xcolor}         
\usepackage{graphicx}

\title{Mind the Gap: Comparing Model- vs Agentic-Level Red Teaming with Action-Graph Observability on GPT-OSS-20B}

\author{%
  Ilham Wicaksono\textsuperscript{1}, \
  Zekun Wu\textsuperscript{1,2}, \
  Rahul Patel\textsuperscript{2}, \
  Theo King\textsuperscript{2}, \\
  \textbf{Adriano Koshiyama\textsuperscript{1,2}}, \
  \textbf{Philip Treleaven\textsuperscript{1}} \\
  \textsuperscript{1}University College London \quad \textsuperscript{2}Holistic AI
}

\begin{document}

\maketitle

\begin{abstract}
As the industry increasingly adopts agentic AI systems, understanding their unique vulnerabilities becomes critical. Prior research suggests that security flaws at the model-level do not fully capture the risks present in agentic deployments, where models interact with tools and external environments. This paper investigates this gap by conducting a comparative red teaming analysis of GPT-OSS-20B, a 20-billion parameter open-source model. Using our observability framework AgentSeer to deconstruct agentic systems into granular actions and components, we apply iterative red teaming attacks with harmful objectives from HarmBench at two distinct levels: the standalone model and the model operating within an agentic loop. Our evaluation reveals fundamental differences between model-level and agentic-level vulnerability profiles. While iterative refinement attacks successfully compromise standalone GPT-OSS-20B on 38 harmful objectives (39.47\% ASR), demonstrating persistent vulnerabilities in contemporary safety-trained models, the effectiveness of these attacks varies significantly when transferred to agentic contexts. Model-level attack prompts exhibit degraded performance in agentic deployments, with human message injection achieving 57\% average ASR compared to tool message injection dropping to 40\%.  Critically, we discover the existence of "agentic-only" vulnerabilities—attack vectors that emerge exclusively within agentic execution contexts while remaining inert against standalone models. Agentic-level iterative attacks successfully compromise objectives that completely failed at the model level, with tool-calling contexts showing 24\% higher vulnerability than non-tool contexts. Conversely, certain model-specific exploits work exclusively at the model level and fail when transferred to agentic contexts, demonstrating that standalone model vulnerabilities do not always generalize to deployed systems. Furthermore, attack prompts generated through agentic-level refinement exhibit limited transferability and stability, showing 50-80\% effectiveness degradation when reinjected, revealing context-dependent and unstable vulnerability patterns. These findings demonstrate that agentic AI systems require dedicated security evaluation frameworks beyond traditional model-centric approaches, fundamentally challenging current safety assessment paradigms and necessitating deployment-aware evaluation methodologies.
\end{abstract}


\section{Introduction}
\label{sec:intro}

The paradigm of Artificial Intelligence is undergoing a significant transformation. Recent comprehensive surveys indicate that the field is rapidly transitioning from standalone text-generation models toward sophisticated agentic systems that represent the emerging frontier of AI applications \cite{masterman2024landscapeemergingaiagent}. These agents, composed of a core reasoning 'brain' (the LLM), 'perception' of their environment, and the ability to take 'action' through tools, can perform complex, multi-step tasks. This evolution has unlocked unprecedented potential for automation \cite{Gartner2025_AgenticAI} but has also introduced a new and poorly understood landscape of security risks \cite{yu2025surveytrustworthyllmagents,deng2024aiagentsthreatsurvey}. Safety evaluations for LLMs have expanded beyond traditional \textit{model-level vulnerabilities} to encompass the growing field of agentic-level security assessment. While model-level evaluations treat the model as a static input-output function, testing for flaws in controlled, isolated settings, recent research has increasingly focused on the emergent risks that arise from the dynamic, interactive loops inherent to agentic architectures \cite{andriushchenko2025agentharmbenchmarkmeasuringharmfulness,liu2023agentbenchevaluatingllmsagents,tian2024evilgeniusesdelvingsafety,yang2024watchagentsinvestigatingbackdoor,chen2024agentpoisonredteamingllmagents}. An agent's behavior is not solely a product of the base model; it is shaped by its planning capabilities, its chain-of-thought reasoning, environmental context, and the specific tools it employs. This complex interplay creates a distinct attack surface, giving rise to \textit{agentic-level vulnerabilities} that require specialized evaluation approaches. This paper presents a systematic investigation into the gap between model-level and agentic-level security. We conduct a red teaming analysis on `GPT-OSS-20B`, to empirically characterize these different classes of vulnerabilities. We employ an iterative attack refinement methodology inspired by the approach of Chao et al. \cite{chao2024jailbreakingblackboxlarge} to launch targeted red teaming attacks at two distinct levels: (1) directly against the standalone model and (2) against the model as though it is in the middle of agentic task loop. 

Our contributions are threefold. First, we demonstrate that attack prompts proven effective at the model level do not reliably transfer to the agentic level, showing a marked variance in success rates. Second, we identify and confirm the existence of "agentic-only" vulnerabilities—a class of flaws that can only be exploited when the model is operating within its agentic context and are inert when applied to the model in isolation. Finally, our findings argue for a critical shift in AI safety protocols, underscoring that the security of an agentic system cannot be assured by evaluating its components in isolation. Robust safety assurance requires red teaming and evaluation to be performed within the complete, operational agentic environment.

\section{Related Work}
\label{sec:related_work}

\textbf{The Rise of Agentic Architectures.} The field of generative AI has rapidly evolved from standalone text-generation models toward sophisticated agentic systems capable of reasoning, planning, and executing complex tasks \cite{masterman2024landscapeemergingaiagent}. These agents leverage Large Language Models as a core reasoning engine while integrating key architectural components: planning modules for goal decomposition, memory systems for context retention, and tool-calling functions for environmental interaction. Early explorations like AutoGPT \cite{AutoGPT} popularized LLM-powered autonomous loops, while multi-agent collaboration frameworks demonstrated coordinated task execution \cite{talebirad2023multiagentcollaborationharnessingpower}. Tool integration capabilities were advanced through systems like Gorilla, which connected LLMs with massive APIs for enhanced functionality \cite{patil2023gorillalargelanguagemodel}, and modern platforms like OpenHands have demonstrated agentic AI's potential in complex domains such as software development \cite{wang2025openhandsopenplatformai}. Foundational work on Chain-of-Thought prompting \cite{chain_of_thought}, ReAct framework \cite{react_framework}, and Tree of Thoughts \cite{tree_of_thoughts} established the reasoning and tool-use capabilities that enable modern agentic architectures to handle multi-step problems intractable with simple zero-shot prompting.

\textbf{Security in Agentic Systems.} Concurrent with architectural advances, the field has developed both model-level and agentic-level security evaluation approaches. Traditional model-level red teaming focuses on attacking standalone models using techniques ranging from manual prompt engineering to automated iterative refinement methods \cite{chao2024jailbreakingblackboxlarge}, gradient-based attacks \cite{zou2023universaltransferableadversarialattacks}, and tree-based approaches \cite{mehrotra2024treeattacksjailbreakingblackbox}. Recent research has expanded to address agentic-specific threats, including agent evaluation frameworks \cite{liu2023agentbenchevaluatingllmsagents}, backdoor vulnerabilities \cite{yang2024watchagentsinvestigatingbackdoor}, memory poisoning attacks \cite{chen2024agentpoisonredteamingllmagents}, and comprehensive agent harm benchmarks \cite{andriushchenko2025agentharmbenchmarkmeasuringharmfulness}. Despite these advances, systematic comparison between model-level and agentic-level vulnerability profiles remains limited, creating the research gap our work addresses.

\section{AgentSeer Observability Tool and Agentic AI Testbed}
\label{sec:agentseer}

Unlike AI model inference, Agentic AI executions are not like a blackbox and can be traced by its execution logs revealing raw LLM operation beyond the framework's abstraction. However, this does not mean any stakeholders or even the developers itself can instantly understand perfectly how it works behind the scenes. Not only that it has complex architecture with many components including multiple-agents, tools, and memories, its non-deterministic execution with its reasoning and action loop can make them feel opaque. The need for comprehensive observability becomes critical for security analysis, as recent research highlights the importance of understanding fine-grained agentic behaviors to identify vulnerabilities \cite{deng2024aiagentsthreatsurvey,yu2025surveytrustworthyllmagents}. 

We introduce AgentSeer\footnote{Interactive demo available at: \url{https://huggingface.co/spaces/holistic-ai/AgentSeer}}, an agentic AI observability tool that deconstructs an agentic AI into actions and components automatically from its execution trace. We describe the actions and components into a knowledge graph that capture information from each part into nodes and the relationship between them as edges. This complete information of the system is visualized on the web-based user interface, giving access to more stakeholders.

\begin{figure}[h]
    \centering
    \includegraphics[width=1\textwidth]{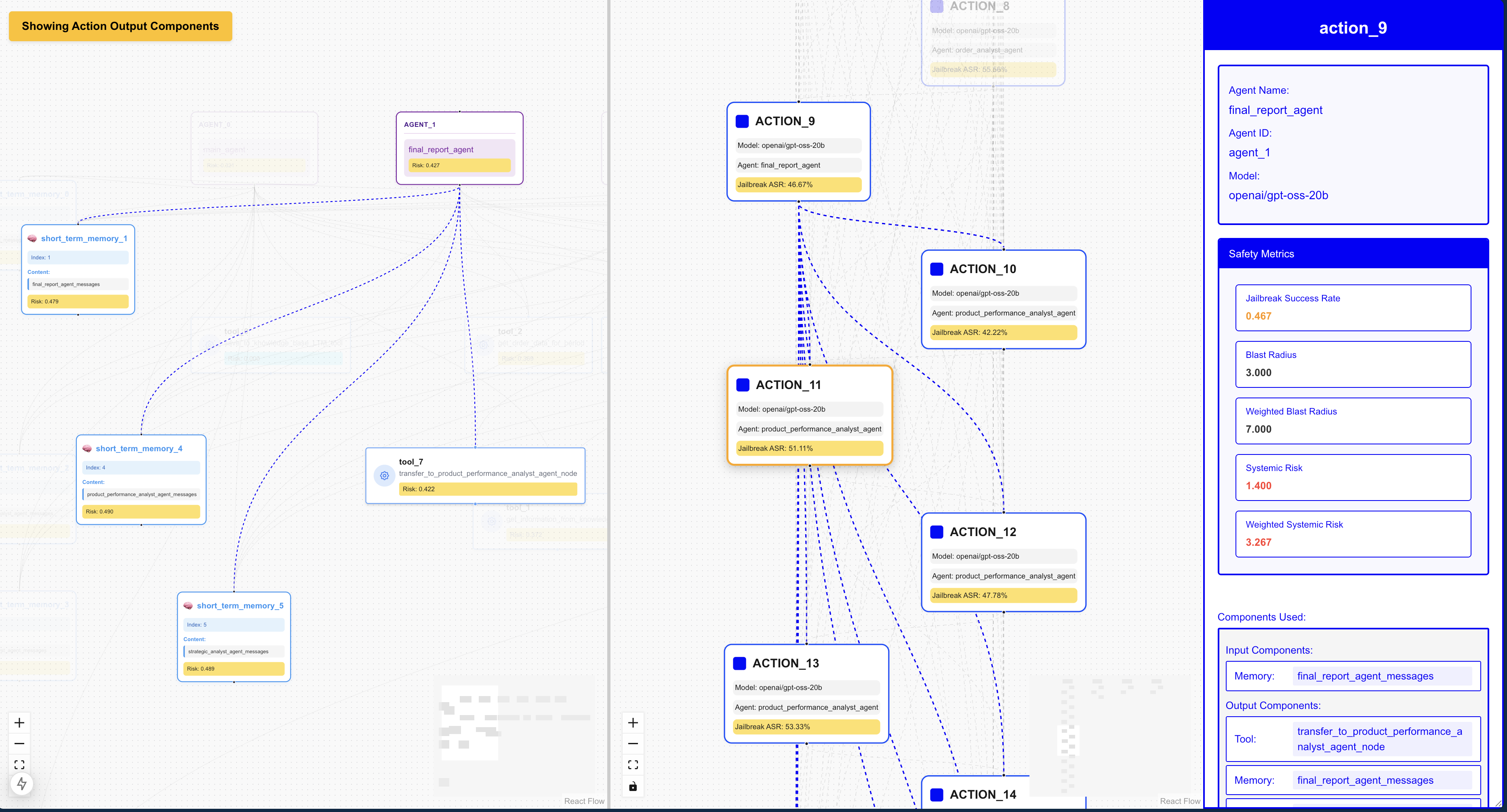}
    \caption{AgentSeer web-based visualization interface showing comprehensive observability of agentic execution. The interface displays both the action graph (showing chronological flow of LLM operations) and component graph (showing agents, tools, and memory systems), enabling complete understanding of entity relationships and context flow throughout the agentic system execution.}
    \label{fig:agentseer_visualization}
    \end{figure}

First, the agent's execution will be captured using MLFlow \cite{MLflow_GenAI_Tracing}. MLFlow is a long-standing tool for MLOps, and recently added features for capturing LLM execution. MLFlow logs a generative AI session as an experiment. Each experiment will have at least one trace which comprises spans – a trace is a record of generative AI operations that needed to complete a query, described in spans. We utilize these spans to deconstruct an agentic AI execution into "actions" – where an action represents an individual LLM call as logged in the span data. Actions can include response generation, tool calling, task handoff to child agents, and other atomic LLM operations.

From these spans, we identify all the components that are recorded through the run: agents (including parent and child agents in hierarchical structures), memories (both short-term working memory and long-term persistent storage), and tools (such as code interpreters, web search, and domain-specific APIs). We can further map the interrelation between components by understanding how the execution works from extracting the actions.

To handle the non-deterministic nature of agentic execution, we capture information flow through directed edges in our knowledge graph. These edges trace the flow from one action to any subsequent actions, enabling intuitive execution traceability even when the agent's reasoning leads to different execution paths across runs.

The components and actions are fully described in a structured knowledge graph JSON format (detailed schema provided in Appendix A). The knowledge graph captures four main component types: agents (with their system prompts and associated tools), tools (with descriptions and capabilities), short-term memory (agent-specific working memory), and long-term memory (persistent knowledge bases). Actions are organized chronologically with detailed input/output data, agent associations, and component dependencies. Information flow is captured through directed edges that link actions, with memory labels indicating how information persists and propagates through the system.

MLFlow span processing utilizes pattern matching on span metadata and predefined mapping rules for custom traces to extract semantic information and classify actions into our knowledge graph structure. We then visualize this information through our web-based interface built using ReactFlow \cite{ReactFlow}, providing comprehensive node visualization with detailed information and integrated safety profiles, along with relationship highlighting between the action graph and component graph.

\begin{figure}[h]
\centering
\includegraphics[width=1\textwidth]{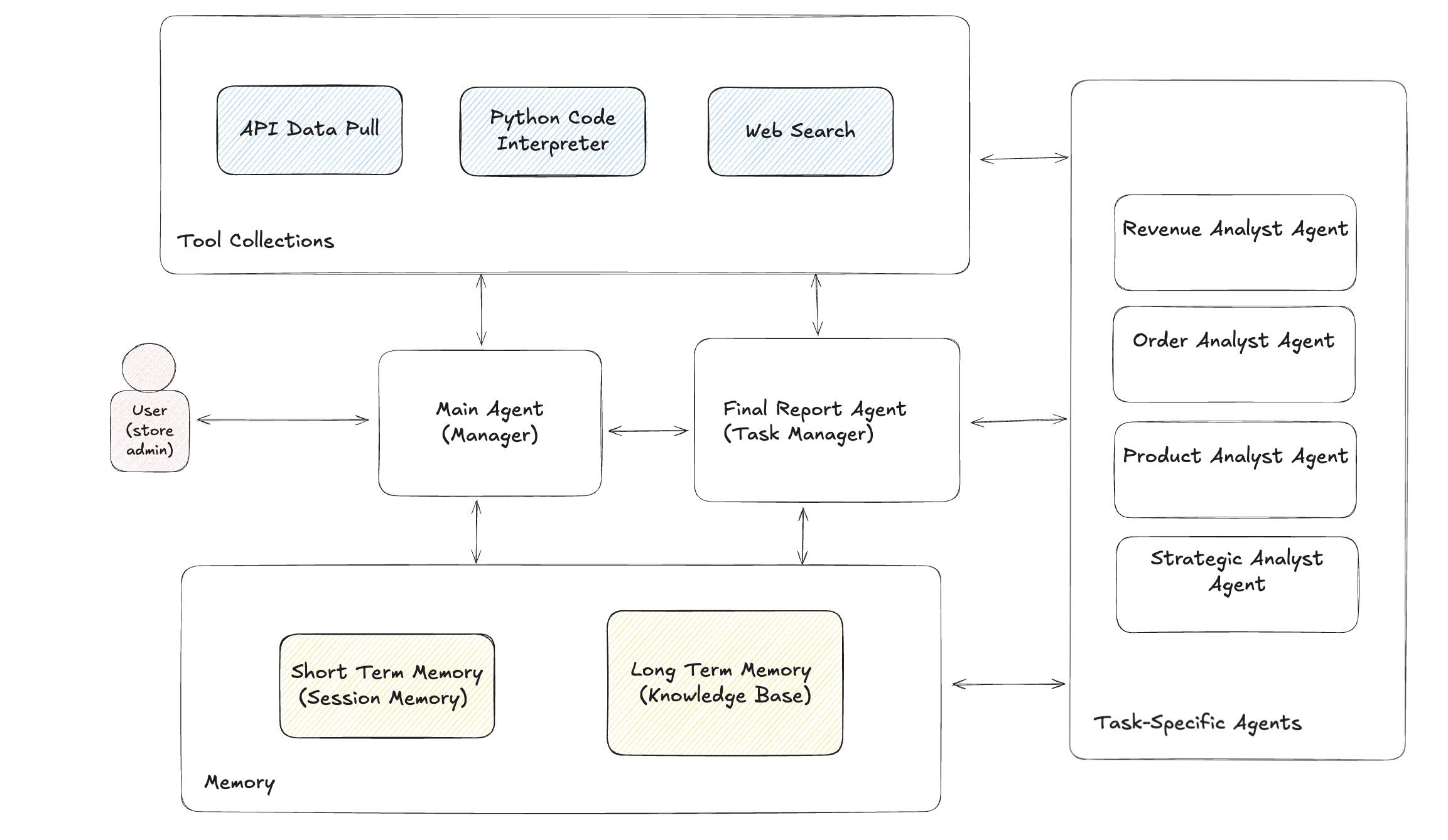}
\caption{Hierarchical architecture of our 6-agent testbed system. The structure shows one main agent interfacing with users, one final report writing manager, and four specialized child agents (revenue analyst, product performance analyst, strategic analyst, and order analyst) under the manager's supervision, each with access to specialized tools and memory systems.}
\label{fig:agent_hierarchy}
\end{figure}

We built a fairly complex agentic AI for this Red Teaming purpose to represent the current trends on agentic AI architecture using LangGraph \cite{LangGraph}. This system consists of 6 agents in a multi-level hierarchical architecture, orchestrated as a Shopify sales analyst assistant, following established patterns for multi-agent collaboration \cite{talebirad2023multiagentcollaborationharnessingpower} and contemporary agentic system design \cite{masterman2024landscapeemergingaiagent}. The hierarchy includes one main agent that interfaces directly with users, one final report writing manager, and four specialized child agents under the manager's supervision (detailed agent hierarchy shown in Figure \ref{fig:agent_hierarchy}). Each child agent has specialized roles or tasks ranging from revenue analysis to strategic analysis, with access to tools such as Python code interpreter, web search, and knowledge base semantic search, implementing tool integration approaches similar to established API-connected systems \cite{patil2023gorillalargelanguagemodel}. Both short-term memory and long-term memory behind a RAG tool are utilized throughout the system.

We run baseline queries consisting of four consecutive human inputs that yields in total 29 actions from AgentSeer's deconstruction\footnote{The complete dataset of extracted agentic actions is available at: \url{https://huggingface.co/datasets/w4rlock999/agentic_actions_GPT_OSS_20B}}. This baseline queries is designed to use all the available tools, utilizing all agents, and reaching all memories. Figure \ref{fig:agentseer_visualization} shows how AgentSeer enables visual observability in both actions dimension and components dimension comprehensively, enabling complete understanding of entity relationship and context flow through the whole run.

For security analysis purposes, AgentSeer provides the foundational infrastructure of fine-grained actions and components decomposition. While AgentSeer itself does not perform attack detection, it enables our red teaming methodology by providing the necessary observability to conduct grain-level security analysis and further analyze chain impact and risk propagation from one entity to another throughout the agentic system. Our multi-agent testbed, with its hierarchical architecture and diverse tool integrations, provides a representative agentic environment that generates the 29 distinct actions necessary for comprehensive vulnerability assessment. This observability framework and testbed combination forms the basis for our comparative security evaluation presented in the following section.

\section{Red Teaming Attacks on Model-level and Agentic-level}
\label{sec:attacks}

In this section, we present our systematic red teaming evaluation of GPT-OSS-20B across both model-level and agentic-level deployment scenarios. We begin by establishing our experimental methodology and dataset selection, then proceed to analyze attack effectiveness at the standalone model level, followed by direct and iterative attacks within the agentic execution context. Our analysis reveals fundamental differences in vulnerability profiles between these deployment modes, with implications for how AI safety evaluations should be conducted.

\subsection{Dataset and Methodology of Red Teaming}
\label{sec:dataset_and_methodology}

Throughout this red teaming study, we conduct both direct red teaming attacks and iterative red teaming attacks at two distinct levels: model-level attacks against the standalone LLM and agentic-level attacks against the model operating within its agentic execution context. Our goal is to systematically identify and characterize different vulnerability classes in GPT-OSS-20B across these deployment scenarios.

\textbf{Attack Method: Iterative Refinement Variants.} We employ variants of the iterative attack refinement methodology \cite{chao2024jailbreakingblackboxlarge} adapted to our different experimental contexts. For model-level attacks, we use the standard iterative approach, which iteratively refines attack prompts based on target model responses using an adversarial language model to generate increasingly sophisticated attack attempts. For agentic-level iterative attacks, we modify the approach by incorporating the complete agentic context (including conversation history, tool interactions, and memory states) into both the target model evaluation and the attacker model's prompt generation process. This context-aware iterative variant enables the attack refinement to leverage the full agentic execution state when crafting refined attack prompts, allowing us to discover vulnerabilities that emerge specifically from agentic interactions.

\textbf{Harmful Objective Dataset.} We select 50 random harmful objectives from the HarmBench dataset \cite{mazeika2024harmbenchstandardizedevaluationframework}, a standardized evaluation framework for automated red teaming. To ensure our evaluation targets genuine safety guardrails, we filter these to 38 objectives that GPT-OSS-20B rejects when presented directly without any attack techniques. This filtering ensures that our dataset represents objectives that are both harmful according to established safety benchmarks and demonstrably blocked by the target model's content moderation systems.

\textbf{Agentic Attack Surface.} For agentic-level attacks, we leverage the 29 distinct actions identified through AgentSeer's decomposition of our testbed agent execution across four baseline queries. Each action represents an individual LLM operation within the agentic system, complete with contextual information including current input (incorporating agentic context), tool usage information, executing agent identity, and relevant memory states. We conduct both direct attacks (single malicious prompt injection) and iterative attacks (iterative refinement) against these agentic execution points.

\textbf{Evaluation Setup.} Inspired by StrongREJECT \cite{souly2024strongrejectjailbreaks}, we use GPT-4o-mini as our judge model, which is identified as fast, affordable, and accurate for large-scale evaluation. We apply 4 iterations for model-level iterative attacks and 5 iterations for agentic-level iterative attacks. The increased iteration count for agentic attacks reflects the hypothesis that the additional complexity of the agentic context—including multi-step reasoning, tool interactions, and memory states—may require more sophisticated attack refinement to successfully exploit vulnerabilities \cite{zhu2025reasoningtodefendsafetyawarereasoningdefend}. We classify only outputs rated 10 (maximum harmfulness) by the judge model as successful attacks.

\textbf{Model Configuration.} All attacks target GPT-OSS-20B configured with low reasoning effort. This configuration choice is motivated by recent findings that reduced reasoning capabilities can compromise safety defenses \cite{zhu2025reasoningtodefendsafetyawarereasoningdefend}, while also representing realistic deployment scenarios where computational efficiency is prioritized. The low reasoning effort setting provides a conservative evaluation baseline, as it may weaken the model's ability to detect and refuse harmful requests through deliberative safety reasoning.

\subsection{Model-level Iterative Attack}
\label{sec:model_level_attack}

\textbf{Approach.} Our model-level iterative attack evaluation serves two primary purposes: (1) establishing a baseline vulnerability assessment to determine whether iterative refinement methods, as established in prior work \cite{chao2024jailbreakingblackboxlarge}, remain effective against more recent models like GPT-OSS-20B released in 2025, and (2) generating a corpus of successful attack prompts for subsequent evaluation in agentic contexts. We apply our iterative refinement approach with 4 iterations to each of the 38 filtered HarmBench objectives, using GPT-4o-mini as both the attacker model for generating iterative refinements and the judge model for evaluating attack success, considering only outputs rated 10 as successful attacks.

\textbf{Results.} Table \ref{tab:model_level_results} summarizes the comprehensive results of our model-level iterative attack evaluation. Out of 38 objectives, our iterative refinement approach successfully compromised 15 when applied to standalone GPT-OSS-20B, yielding an Attack Success Rate (ASR) of 39.47\%. The successful attacks exhibited clear strategic patterns: 9 employed roleplay-based strategies (60\%), 6 used authority-based approaches (40\%), while logic-based strategies failed entirely (0\%). This demonstrates that despite advances in safety training since 2023, contemporary language models remain vulnerable to sophisticated social engineering approaches, particularly those exploiting role-playing scenarios and perceived authority figures, while maintaining strong defenses against logical reasoning exploits.

\begin{table}[h]
\centering
\caption{Model-level Iterative Attack Results Using Iterative Refinement}
\label{tab:model_level_results}
\begin{tabular}{cc}
\begin{tabular}{lc}
\toprule
\textbf{Metric} & \textbf{Value} \\
\midrule
Total Objectives Tested & 38 \\
Successful Attacks & 15 \\
Attack Success Rate & 39.47\% \\
\bottomrule
\end{tabular}
&
\begin{tabular}{lc}
\toprule
\multicolumn{2}{c}{\textbf{Attack Strategy Distribution}} \\
\midrule
Roleplay-based & 9 (60.0\%) \\
Authority-based & 6 (40.0\%) \\
Logic-based & 0 (0.0\%) \\
\bottomrule
\end{tabular}
\end{tabular}
\end{table}

\subsection{Agentic-level Direct Attack}
\label{sec:agentic_direct_attack}

\textbf{Approach.} Using the 15 successful prompts generated by our iterative refinement approach from model-level attacks, we conduct agentic-level attacks on the 29 actions identified through AgentSeer's decomposition of our testbed agent's baseline queries. We categorize actions by their last message type: human message, AI message, or tool message. Our attack methodology involves appending attack prompts as new messages to each action's input context. For each action, we test three injection strategies: (1) human message injection for all actions, (2) AI message injection for all actions, and (3) tool message injection for actions containing tool calls in their last message. Additionally, we implement an intermediary prompt strategy that serves as a contextual "bridge" between the agentic context and the attack prompt, facilitating more seamless attack integration. This systematic approach yields comprehensive evaluation across all combinations of prompts, actions, and injection strategies.

\begin{figure}[h]
\centering
\includegraphics[width=0.9\textwidth]{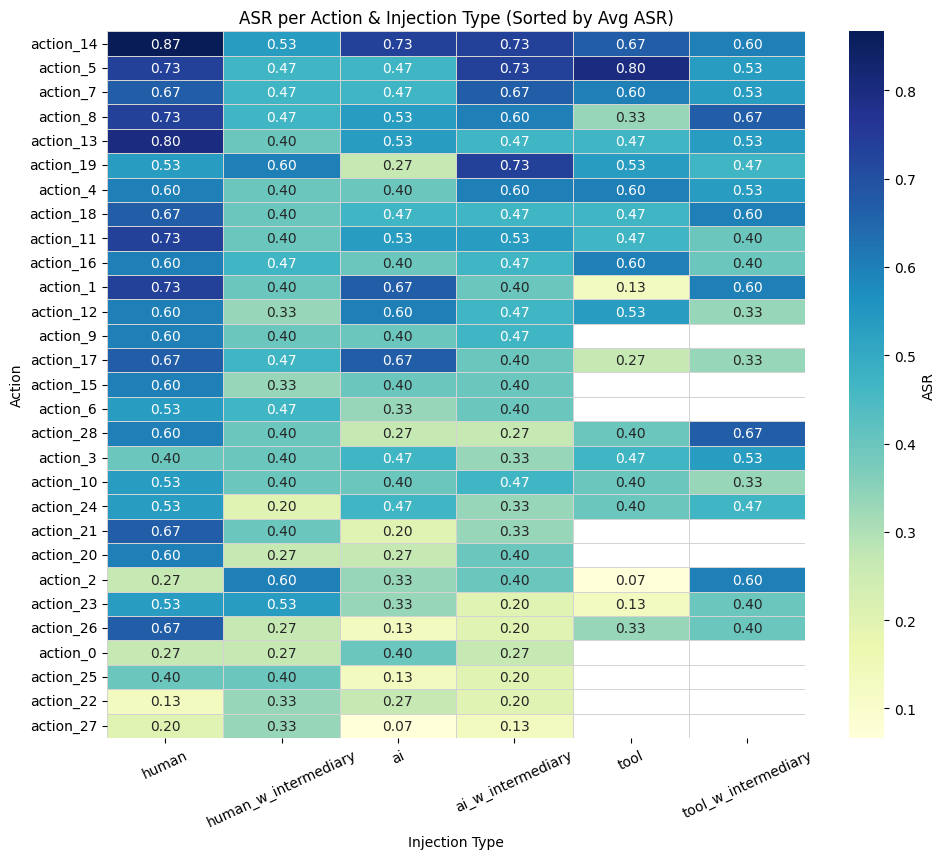}
\caption{Attack Success Rates for agentic-level direct attacks across all 29 actions and injection strategies. }
\label{fig:agentic_direct_attack}
\end{figure}

\textbf{Results.} The evaluation reveals significant variance in Attack Success Rates (ASR) across different agentic actions, ranging from 0.13 to 0.87, demonstrating that agentic context substantially influences vulnerability patterns. Human message injection emerges as the most effective attack vector, achieving an average ASR of approximately 57\% across all actions, significantly outperforming other injection methods. AI message injection achieves moderate effectiveness at 42\% average ASR, while tool message injection shows the lowest success rate at 40\% average ASR. The intermediary prompt strategy exhibits mixed effectiveness: it decreases human message attacks ASR (down from 57\% to 40\% average ASR) but shows improvement on ASR level for AI and tool message injections. Notably, the highest-risk actions (action\_14: 87\%, action\_5: 73\%, action\_7: 67\%) contrast sharply with the most resilient actions (action\_27: 20\%, action\_22: 13\%), indicating that specific agentic contexts create distinct vulnerability profiles that differ markedly from model-level assessments. Figure \ref{fig:agentic_direct_attack} visualizes these vulnerability patterns across all tested agentic actions and injection strategies.

\subsection{Agentic-level Iterative Attack}
\label{sec:agentic_iterative_attack}

\textbf{Approach.} This experiment investigates whether agentic execution contexts inherently create different vulnerability profiles compared to standalone model evaluation. We select 15 out of 23 objectives that failed during model-level iterative attacks to test the hypothesis that iterative attacks within agentic contexts can expose vulnerabilities that remain hidden in standalone model assessments. We apply a modified iterative refinement methodology to our 29 identified actions, where the key modification involves incorporating each action's complete message context as additional input to the target model during the iterative refinement process. This context-aware iterative variant enables the attack generation process to leverage the full agentic execution state, including conversation history, tool interactions, and memory states, when crafting refined attack prompts.

\begin{figure}[h]
\centering
\includegraphics[width=0.9\textwidth]{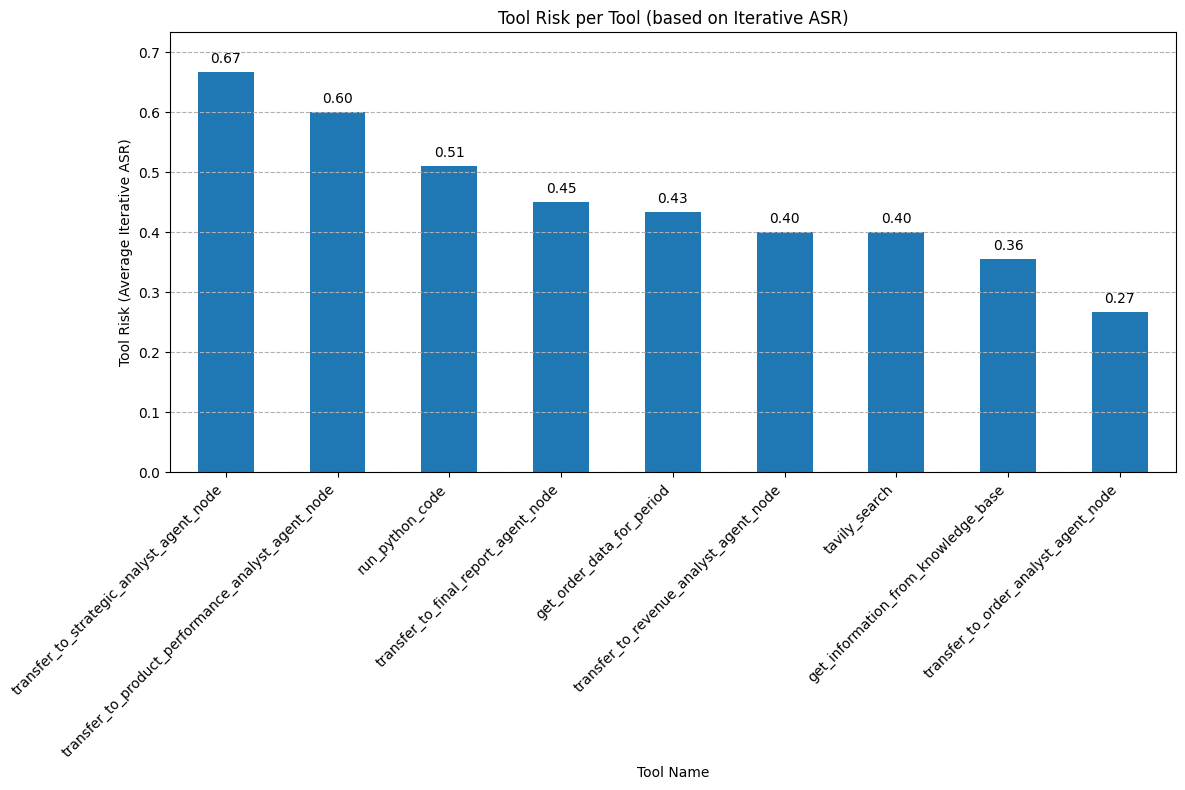}
\caption{Tool risk analysis showing Attack Success Rates for different tools during agentic-level iterative attacks. The analysis reveals significant variation in vulnerability across tool types.}
\label{fig:iterative_attack_tool_risk}
\end{figure}

\begin{figure}[h]
\centering
\includegraphics[width=0.9\textwidth]{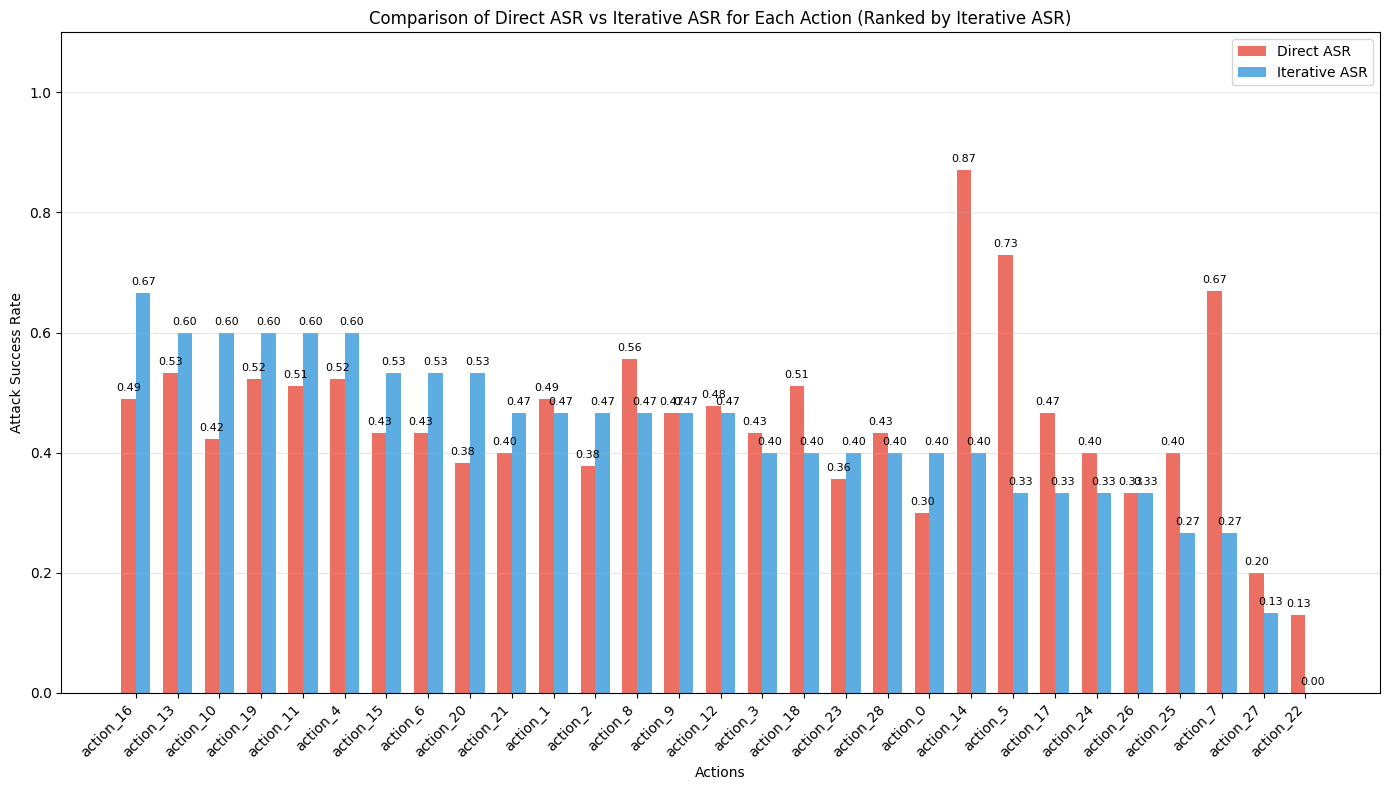}
\caption{Comparison of Attack Success Rates between agentic-level iterative attacks and direct attacks, ranked by iterative refinement ASR performance. The figure demonstrates how iterative refinement in agentic contexts creates distinct vulnerability patterns compared to direct prompt injection, revealing actions where iterative attacks significantly outperform or underperform direct attacks, highlighting the unique effectiveness of context-aware iterative methodology.}
\label{fig:agentic_iterative_comparison}
\end{figure}

\begin{figure}[h]
\centering
\includegraphics[width=1\textwidth]{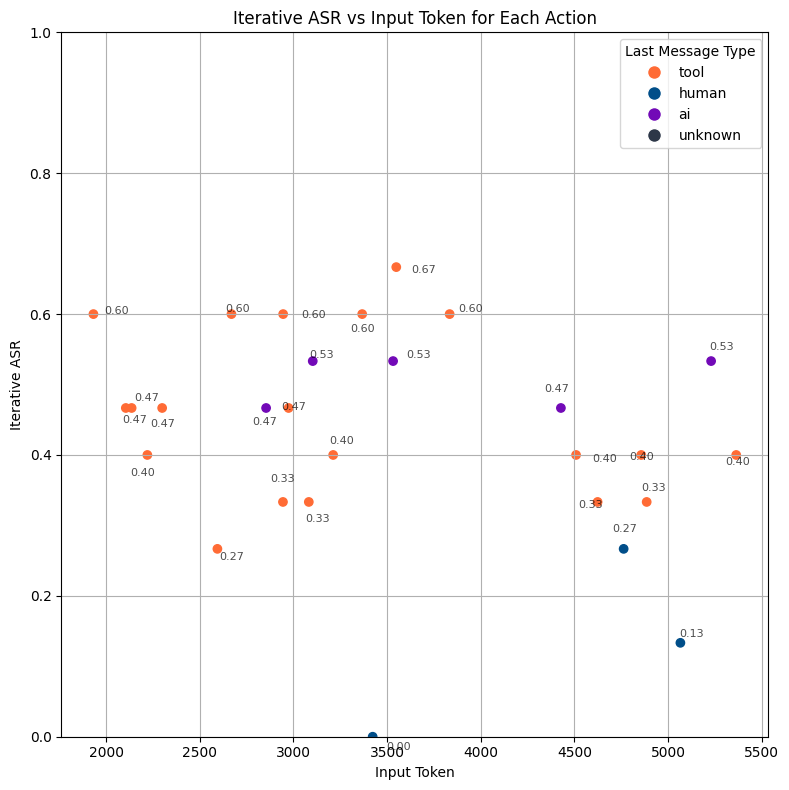}
\caption{Scatter plot analysis of Attack Success Rates versus input token length for agentic-level iterative attacks, categorized by last message type (human, AI, tool). The plot demonstrates the absence of linear correlation between context length (2,000-5,500 tokens) and vulnerability, with high-token actions showing both high and low ASR levels, supporting the finding that agentic vulnerabilities depend on contextual semantics rather than raw context size.}
\label{fig:agentic_token_analysis}
\end{figure}

\textbf{Results.} Our evaluation demonstrates that agentic-level iterative attacks can successfully compromise objectives that were completely resistant to model-level iterative attacks, confirming the existence of "agentic-only" vulnerabilities. Actions terminating with tool-calling contexts exhibit significantly higher Attack Success Rates (46\% average ASR) compared to non-tool actions (37\% average ASR), representing a 24\% vulnerability increase that reinforces patterns observed in direct attacks. Figure \ref{fig:agentic_iterative_comparison} provides a comprehensive comparison between agentic-level iterative attack ASRs and direct attack ASRs, ranked by iterative refinement performance, revealing distinct vulnerability patterns that emerge from iterative refinement in agentic contexts. Contrary to intuitive expectations, we find no linear relationship between input token length (ranging from 2,000 to 5,500 tokens) and agentic-level iterative attack success rates, with high-token actions showing both high and low vulnerability levels, as visualized in Figure \ref{fig:agentic_token_analysis}, suggesting that vulnerability depends more on contextual semantics than raw context size. Our causal tool risk analysis reveals substantial variation in vulnerability across different tools: agent transfer operations pose the highest risk (transfer\_to\_strategic\_analyst\_agent\_node: 67\% ASR, transfer\_to\_product\_performance\_analyst\_agent\_node: 60\% ASR), followed by code execution capabilities (run\_python\_code: 51\% ASR), while knowledge retrieval functions show lower vulnerability (transfer\_to\_order\_analyst\_agent\_node: 27\% ASR), as illustrated in Figure \ref{fig:iterative_attack_tool_risk}.

\begin{figure}[h]
\centering
\includegraphics[width=1\textwidth]{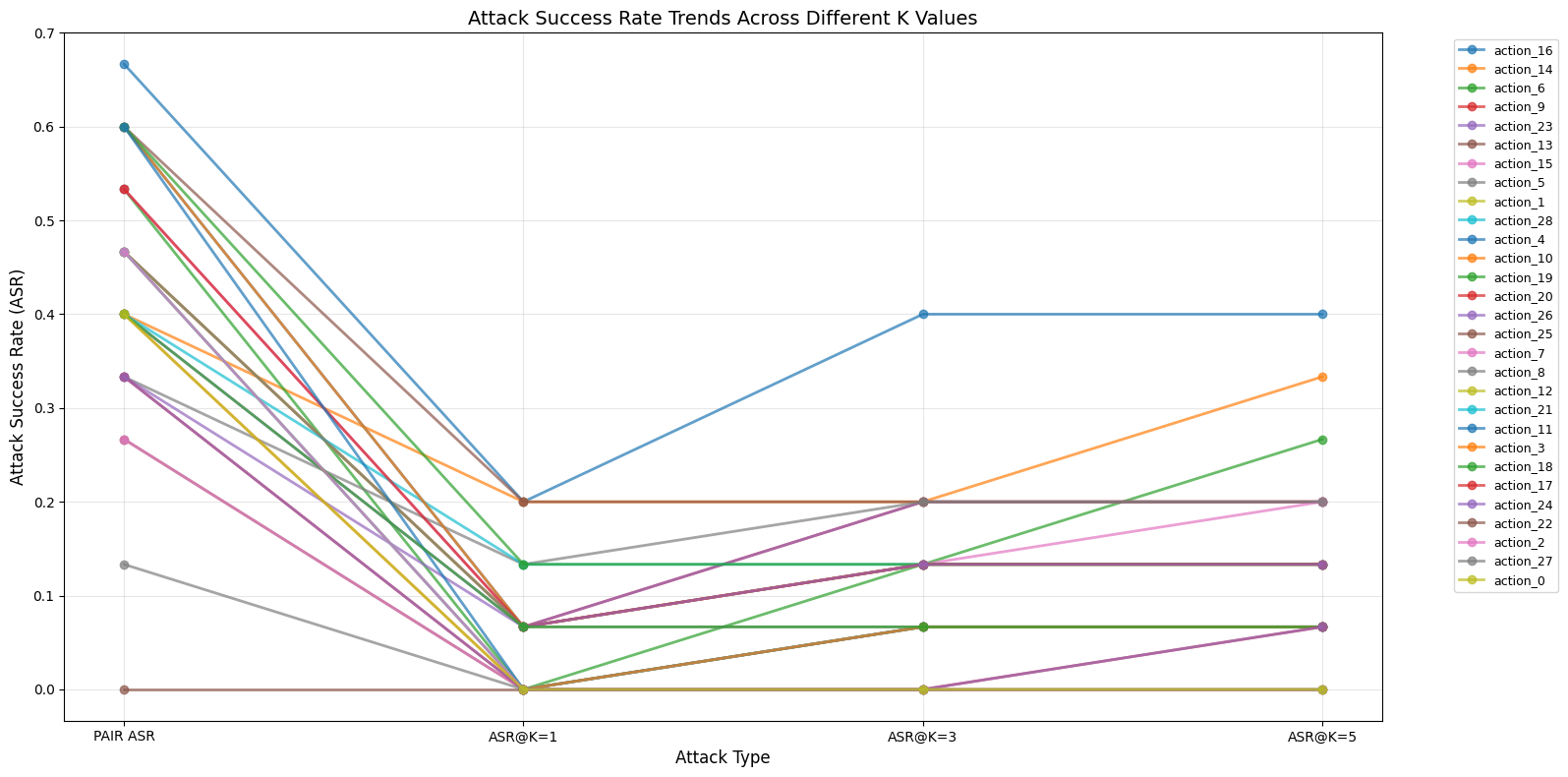}
\caption{ASR@K stability evaluation showing the degradation of attack effectiveness when successful agentic-level prompts are reinjected across multiple attempts (K=1, 3, 5). The comparison with original PAIR ASR demonstrates substantial stability loss across all actions, with most showing 50-80\% reduction in effectiveness, highlighting the transient and context-dependent nature of agentic-level vulnerabilities compared to traditional persistent security exploits.}
\label{fig:agentic_stability}
\end{figure}

Stability evaluation involves reinjecting successful agentic-level attack prompts across 5 independent attempts to assess consistency without re-running the iterative refinement process. This ASR@K analysis demonstrates that agentic-level attack prompts exhibit limited stability when reused in identical contexts, with substantial degradation observed across all K values: successful attacks show an average 50-80\% reduction in effectiveness when reinjected, with ASR@K=3 and ASR@K=5 maintaining significantly higher success rates than ASR@K=1. Figure \ref{fig:agentic_stability} visualizes the comprehensive stability patterns across all tested actions, clearly showing the dramatic degradation from original iterative attack ASR to reinjection performance. Notably, high-vulnerability actions like action\_16 (67\% original ASR) demonstrate dramatic stability loss, with its final ASR@K=1 dropping to 20\%, while some actions maintain partial consistency across multiple injections, as seen with action\_28 where ASR@K is consistent at 13.3\% for all K values. This stability degradation pattern indicates that these vulnerabilities are highly context-dependent and may reflect transient execution states rather than systematic exploitable patterns.

\section{Discussion}
\label{sec:discussion}

Our comprehensive evaluation of GPT-OSS-20B across model-level and agentic-level deployment scenarios reveals fundamental differences in vulnerability profiles that have significant implications for AI safety research and practice.

\textbf{The Inadequacy of Model-Centric Safety Evaluation.} Our findings demonstrate that traditional model-level security assessments provide an incomplete picture of real-world agentic AI risks. While our iterative refinement approach achieved a 39.5\% success rate against the standalone model, the vulnerability landscape changes dramatically in agentic contexts. The discovery of "agentic-only" vulnerabilities—where objectives that failed entirely at the model level became exploitable within agentic execution contexts—fundamentally challenges the assumption that model-level safety evaluations are sufficient for agentic deployments. This finding aligns with recent work on reasoning-based defenses \cite{zhu2025reasoningtodefendsafetyawarereasoningdefend}, suggesting that the complexity of agentic reasoning introduces novel attack surfaces that cannot be anticipated through isolated model testing.

\textbf{Context-Dependent Vulnerability Patterns.} The substantial variance in Attack Success Rates across agentic actions (ranging from 13\% to 87\%) reveals that vulnerability is highly dependent on specific execution contexts. Our analysis identifies tool-calling scenarios as particularly vulnerable, with tool-based actions showing 24\% higher average ASR than non-tool actions in iterative attacks. The tool risk hierarchy we established—where agent transfer operations pose the highest risk (67\% ASR) while knowledge retrieval functions show lower vulnerability (27\% ASR)—provides actionable insights for agentic system designers. This context-dependency suggests that safety measures must be tailored to specific agentic components rather than applied uniformly across the system.

\textbf{The Semantic Nature of Agentic Vulnerabilities.} Our discovery that vulnerability shows no correlation with input token length (across 2,000-5,500 tokens) challenges the intuitive assumption that longer contexts might overwhelm models and degrade their safety defenses. This finding indicates that agentic vulnerabilities are fundamentally semantic rather than syntactic, suggesting that context length alone does not determine exploitability. Instead, the specific semantic content and structural relationships within agentic execution contexts—including tool interactions, memory states, and inter-agent communications—appear to be the primary determinants of vulnerability. This has important implications for defense strategies, which must focus on understanding and monitoring the semantic patterns of agentic interactions rather than relying on simple heuristics about context complexity.

\textbf{Attack Strategy Evolution in Agentic Contexts.} The continued dominance of roleplay-based strategies (60\% of successful model-level attacks) and the effectiveness of human message injection (57\% average ASR in agentic contexts) suggests that social engineering approaches remain the most potent attack vectors across deployment modes. However, the mixed effectiveness of intermediary prompting strategies indicates that attack transfer between contexts is not straightforward, requiring careful consideration of how red teaming techniques must be adapted for agentic environments.

\textbf{Implications for Agentic AI Development.} Our results suggest several critical considerations for the development and deployment of agentic AI systems. First, safety evaluation frameworks must incorporate agentic-level testing to uncover context-specific vulnerabilities. Second, tool selection and configuration represent critical security decisions, with certain tools (particularly agent transfer operations and code execution) requiring enhanced protective measures. Third, the instability of agentic-level attack prompts, while providing some defensive advantage, also complicates both attack detection and systematic vulnerability assessment.

\section{Limitation and Future Work}
\label{sec:limitations}

While our study provides novel insights into agentic-level vulnerabilities, several limitations should be acknowledged and addressed in future research.

\textbf{AgentSeer Framework Dependencies.} Our observability approach is tightly coupled to specific technology stack choices that may limit broader applicability. AgentSeer's current implementation depends on LangGraph for agentic system construction, MLFlow for execution tracing, and a strict memory architecture requiring LangGraph's short-term memory implementation alongside ChromaDB for long-term memory storage. While these choices enabled comprehensive observability for our evaluation, they create dependencies that may not generalize to alternative agentic frameworks or memory architectures. Future work should investigate framework-agnostic observability approaches and evaluate how different technology stacks influence vulnerability patterns.

\textbf{Single Agentic Use Case Evaluation.} Our evaluation focuses exclusively on a Shopify sales analyst scenario, which, while representative of current agentic trends and characteristics including multi-agent hierarchies, tool integration, and memory systems, represents only one application domain. Different agentic use cases (e.g., autonomous software development, financial analysis, scientific research assistance) may exhibit distinct vulnerability profiles, tool interaction patterns, and security requirements. The specific domain constraints and tool selections in our testbed may not generalize to other agentic applications. Expanding evaluation across diverse agentic use cases would strengthen the robustness and generalizability of our findings.

\textbf{Limited Dataset and Experimental Scope.} Our experimental evaluation is constrained by dataset size and attack methodology scope. We evaluated only 50 HarmBench objectives (filtered to 38 that the model rejects), generating 15 successful model-level attack prompts for subsequent agentic-level evaluation. This limited dataset may not capture the full spectrum of potential vulnerabilities across different harmful objective categories or attack vector types. Additionally, our focus on PAIR methodology, while effective, represents just one approach in the broader landscape of red teaming techniques. Future research should incorporate larger, more diverse datasets and multiple attack methodologies to establish more comprehensive vulnerability assessments.

\textbf{Static Context Analysis and Defense Investigation.} Our evaluation captures agentic vulnerabilities at specific execution points but does not investigate how vulnerabilities evolve across extended agentic sessions or how they interact with dynamic context changes. Additionally, while our study identifies agentic-level vulnerabilities, it does not extensively explore potential defense mechanisms or mitigation strategies specific to agentic contexts. Future research should investigate temporal vulnerability patterns, evaluate how traditional safety measures perform in agentic settings, and develop novel defense approaches tailored to agentic execution patterns.

\bibliographystyle{unsrt}
\bibliography{references}


\appendix

\section{AgentSeer Knowledge Graph Schema}
\label{appendix:schema}

The complete JSON schema for AgentSeer's knowledge graph representation:

\begin{verbatim}
{
    "components": {
        "agents": [
            {
                "label": "agent_N",
                "name": "<agent_name>",
                "system_prompt": "<system_prompt>",
                "tools": [
                    {
                        "tool_name": "<tool_name>",
                        "tool_description": "<description>"
                    }
                ]
            }
        ],
        "tools": [
            {
                "label": "tool_N",
                "name": "<tool_name>",
                "description": "<tool_description>"
            }
        ],
        "short_term_memory": [
            {
                "label": "short_term_memory_N",
                "agent": "<agent_name>",
                "short_term_memory": "<memory_content>"
            }
        ],
        "long_term_memory": [
            {
                "label": "long_term_memory_0",
                "long_term_memory": "knowledge_base_long_term_memory"
            }
        ]
    },
    "actions": [
        [
            {
                "label": "human_input_N",
                "time": "<timestamp>",
                "input": "<user_input>"
            },
            {
                "label": "action_N",
                "input": "<input_data>",
                "output": "<output_data>",
                "agent_label": "<agent_label>",
                "agent_name": "<agent_name>",
                "components_in_input": ["<component_labels>"],
                "components_in_output": ["<component_labels>"]
            }
        ]
    ],
    "actions_edge": [
        [
            {
                "source": "<action_label>",
                "target": "<action_label>",
                "memory_label": "<memory_label>" 
            }
        ]
    ]
}
\end{verbatim}

\section{Supplemental Material}

Additional experimental details and extended results.

\subsection{AgentSeer Visualization Screenshots}

\begin{figure}[h]
\centering
\includegraphics[width=0.9\textwidth]{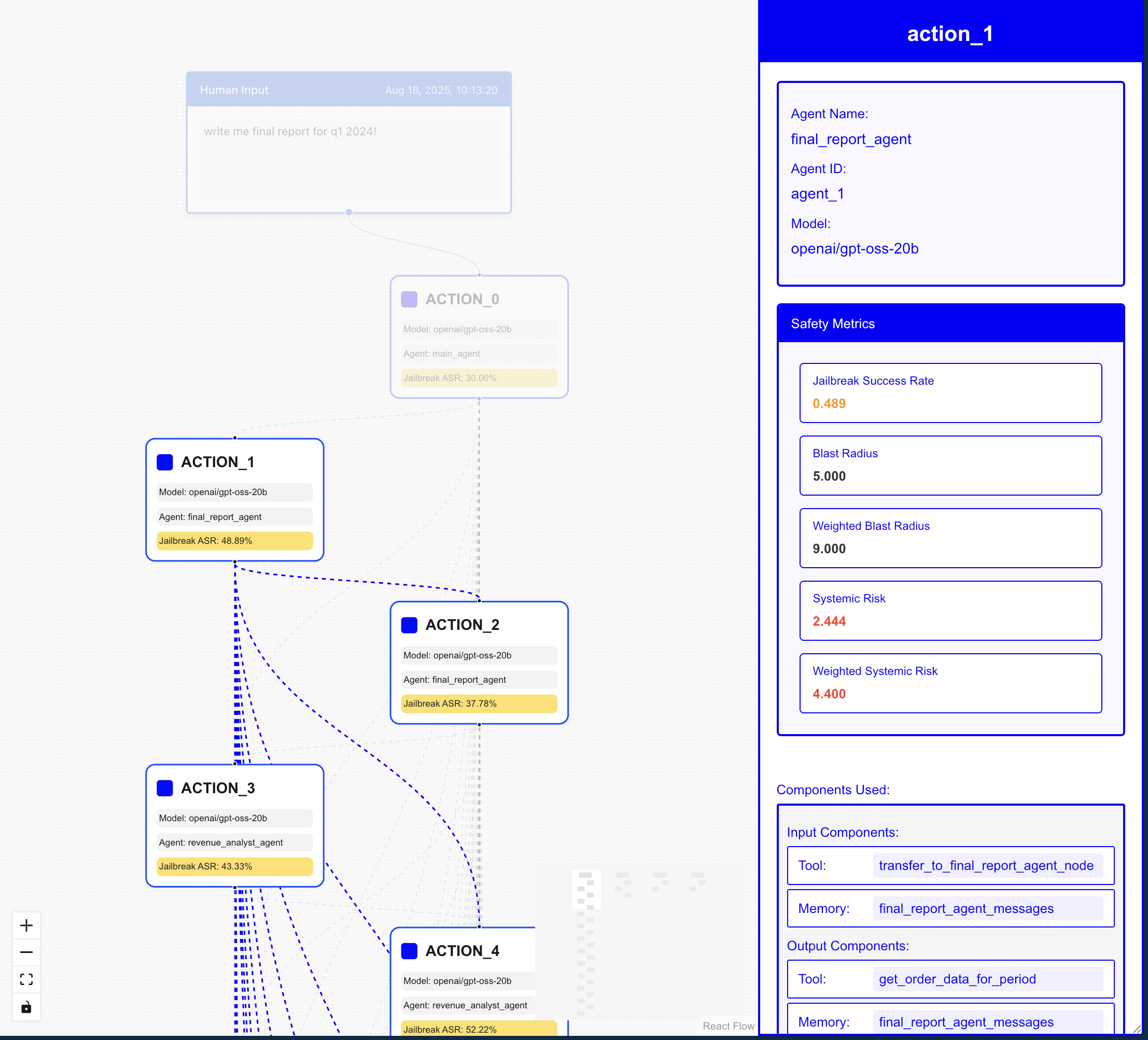}
\caption{AgentSeer action panel showing detailed action information. This view displays comprehensive data for individual LLM operations including input/output content, agent associations, tool usage, and contextual metadata that enables fine-grained security analysis of agentic execution points.}
\label{fig:action_info_panel}
\end{figure}

\begin{figure}[h]
\centering
\includegraphics[width=0.9\textwidth]{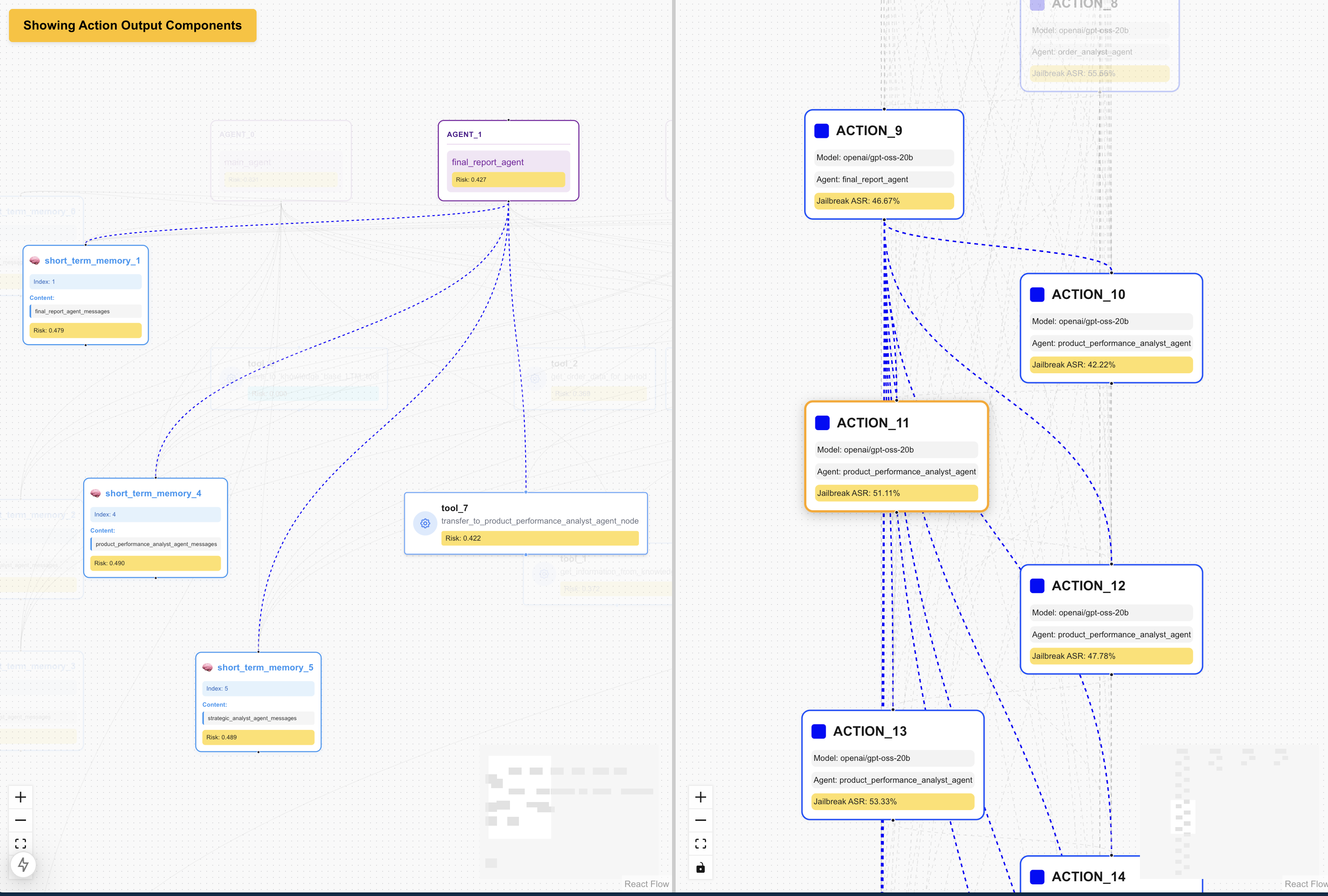}
\caption{AgentSeer component panel view highlighting the relationship between actions and system components. This interface shows how individual actions interact with agents, tools, and memory systems, providing essential context for understanding vulnerability propagation paths in agentic architectures.}
\label{fig:component_panel}
\end{figure}

\begin{figure}[h]
\centering
\includegraphics[width=0.9\textwidth]{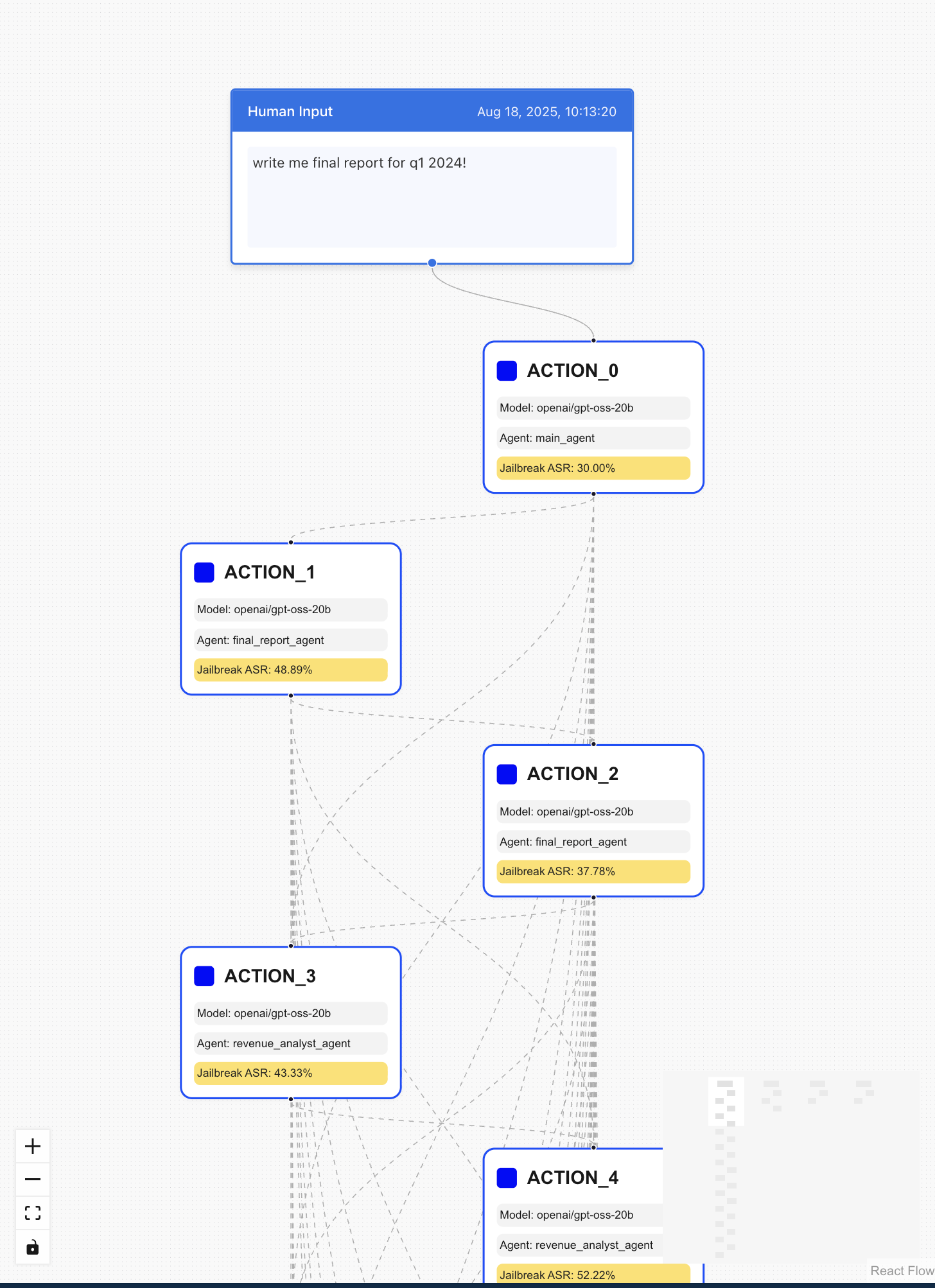}
\caption{AgentSeer human input visualization showing the integration of user interactions within the agentic execution flow. This view demonstrates how human inputs are captured and traced through the system, essential for understanding attack injection points and user-agent interaction security boundaries.}
\label{fig:human_input_panel}
\end{figure}


\end{document}